\documentclass[letterpaper]{article}
\usepackage{iccc}
\usepackage{url}

\usepackage{times}
\usepackage{helvet}
\usepackage{courier}
\usepackage{graphicx}

\usepackage{caption}
\usepackage{subcaption}
\usepackage{hyperref}

\usepackage{balance}

\usepackage{array}
\newcolumntype{P}[1]{>{\centering\arraybackslash}p{#1}}
\newcolumntype{M}[1]{>{\centering\arraybackslash}m{#1}}
\newcolumntype{N}[1]{>{\arraybackslash}m{#1}}

 \title{Creative Robot Dance with Variational Encoder}
\author{A. Augello, E. Cipolla, I. Infantino, A. Manfr\`e, G. Pilato, F. Vella \\
Institute for High Performance Computing and Networking\\
National Research Council of Italy (ICAR-CNR)\\
Via Ugo La Malfa 153\\
Palermo, 90146 Italy\\
}
\begin{document}

\maketitle

\begin{abstract}
What we appreciate in dance is the ability of people to spontaneously improvise new movements and choreographies, surrendering to the music rhythm, being inspired by the current perceptions and sensations and by previous experiences, deeply stored in their memory.
Like other human abilities, this, of course, is challenging to reproduce in an artificial entity such as a robot. Recent generations of anthropomorphic robots, the so-called humanoids, however, exhibit more and more sophisticated skills and raised the interest in robotic communities to design and experiment systems devoted to automatic dance generation.
In this work, we highlight the importance to model a computational creativity behavior in dancing robots to avoid a mere execution of preprogrammed dances. In particular, we exploit a deep learning approach that allows a robot to generate in real time new dancing movements according to to the listened music.
\end{abstract}


\section*{Introduction}
The execution of artistic acts is certainly one of the most fascinating and impactful human activity that robotics aims to replicate.
The abilities of new generation robots can be thoroughly tested in artistic domains, such as dance, music, painting and drama.

The appearance and the skills that characterize anthropomorphic robots make the dance domain very interesting and challenging since human movements can either be replicated, albeit imperfectly, or adapted to the embodiment of the robot. Dance is a harmonic composition of movements driven by the music stimuli and by the interactions with other subjects. Dancing movements follow the rhythm of the music and are synchronous with the song progression. Therefore both the timing and rhythm of the execution of the movements must be taken into account while trying to imitate human behavior.

The implementation of dancing capabilities in robots is not purely pursued for entertainment purposes. It provides new clues to deepen and improve various research thematics, because it requires a robust learning phase that involves both a real-time analysis of music, the choice of harmonious and suitable movements, and moreover social behavior and interaction capabilities \cite{aucouturier2008cheek,augello2016creation,shinozaki2007concept}.

The challenge lies in going beyond a preprogrammed dance executed by robots. A creative process should model the mental processes involved in human creativity to generate movements and taking into account different music genres. The robots' perceptions should influence the choice processes and output of a learning process should lead to conceive a personal artistic style, that could be reconsidered or refined after the audience evaluation. 


While some interesting and creative approach has been proposed for the generation of movements and choreographies \cite{jacob2015viewpoints,carlson2016cochoreo,crnkovic2016generative,lapointe2005dancing}, few are the works aimed at injecting a computational creativity behavior in dancing humanoids \cite{vircikova2010dance,augello2016creation,manfre2016automatic}.

In this work we explore the possibility for a robot to improvise a choreography, building on actions that are either stored in a memory or derived from a continuous elaboration of previous experiences. 
In our proposal, we took inspiration from human dance to create a dataset of movements that the robot can employ in his dance. The dataset, including also information about music features related to the sequences of movements, is used to train a variational encoder. This network allows obtaining a variation of the learned movements according to the listened music. 
The resulting movements are new but are coherent with the learned ones and are well synchronized with the listened music.

\section{State of the art}

Different works in literature propose approaches for dancing motions generation. One of these has been proposed and experimented by Luka and Louise Crnkovic-Friis \cite{crnkovic2016generative}. It is a deep learning generative model, exploiting a Long Short-Term Memory type of recurrent neural network, that is used to produce novel choreographies. The network is trained on raw motion capture data, consisting of contemporary dance motions performed by a choreographer. The training dataset does not contain any information about musical features.

Cochoreo \cite{carlson2016cochoreo} is a module of a sketching tool, named danceForms, used to create and animate keyframes. The module combines the functionality of a creativity support tool with an autonomously creative system, relying on a genetic algorithm, which generates novel keyframes for body positions. The new keyframes are evaluated according to a parameterized fitness function that allows the choreographer to set generation options based on their personal preferences. 

Another example is the work proposed in \cite{aucouturier2007making}, exploiting chaotic itinerancy (CI) dynamics generated by a network of artificial spiking neurons. The motions are chosen in real-time by converting the output of a neural network that processes the musical beats; then are executed by a vehicle-like robot.

For what concerns performances executed by humanoids, generally the interest is in the coordination of dancing gestures and postures according to the detected beats \cite{ellenberg2008exploring,grunberg2009creating,seo2013autonomous,shinozaki2007concept}

Some creative approaches are discussed in \cite{augello2016creation,infantino2016robodanza,manfre2016exploiting,vircikova2010dance,eaton2013approach,xia2012autonomous}.
  Zhou et al. have analyzed some of these works in the taxonomy of robotic dance systems discussed in \cite{peng2015robotic}. In addition to the already mentioned chaotic dynamic  \cite{aucouturier2007making}, they  describe other approaches that have been proposed for the generation of  dance choreographies.  Meng et al. \cite{meng2014robots}, propose the use of the interactive reinforcement learning (IRL), to make robots learn to dance according to human preferences. Among the evolutionary computing approaches Zhou et al. \cite{peng2015robotic}, cite the algorithms proposed in \cite{eaton2013approach,vircikova2010dance}. In detail, the authors of \cite{vircikova2010dance} have initialized a population of individuals encoded from dance characteristics.  Then, the value of fitness of the algorithm is obtained with the interaction of the system's users, obtaining dances reflecting personal preferences. Another approach exploits the Markov Chain Model. As an example, in  \cite{xia2012autonomous}, each motion is considered as a  Markov chain state, and the next motion is determined by the previous motion and the current music emotion.
In ours previous works \cite{infantino2016robodanza,augello2016creation,manfre2016automatic} we proposed the use of both evolutionary computing and Markov models.
We described the system underlying a humanoid dancing performance called ROBODANZA, also discussing the impact on different types of audience. In the performance, a humanoid robot interacts and dances with professional dancers, autonomously following the rhythms suggested by the dancers clapping the hands and tap on a table.
The movements of the robot are generated according to a Hidden Markov model. Different emission matrices determine different execution styles of dance. The Transition Matrix (TM) of the HMM takes into account how a movement follows the previous one in a sequence of dance. It derives from observing the composition and occurrences of the human movement. The creative computational process exploits an interactive genetic algorithm that make the EMs to evolve according to the final user evaluations. Therefore each performance is always different.


\section{Variational Autoencoder}
\label{sect:VE}
Autoencoders have been enjoying significant interest as they can perform a lossy data compression starting from a specific dataset. Once trained, they represent in the hidden layer all the data have been previously exposed to; the representation is ``lossy'' since the reconstructed $x$ is not perfectly identical to the original one, this difference being determined by the chosen distance or ``loss'' function.

\begin{figure}[htb]
\begin{center}
\includegraphics[width=0.45\textwidth]{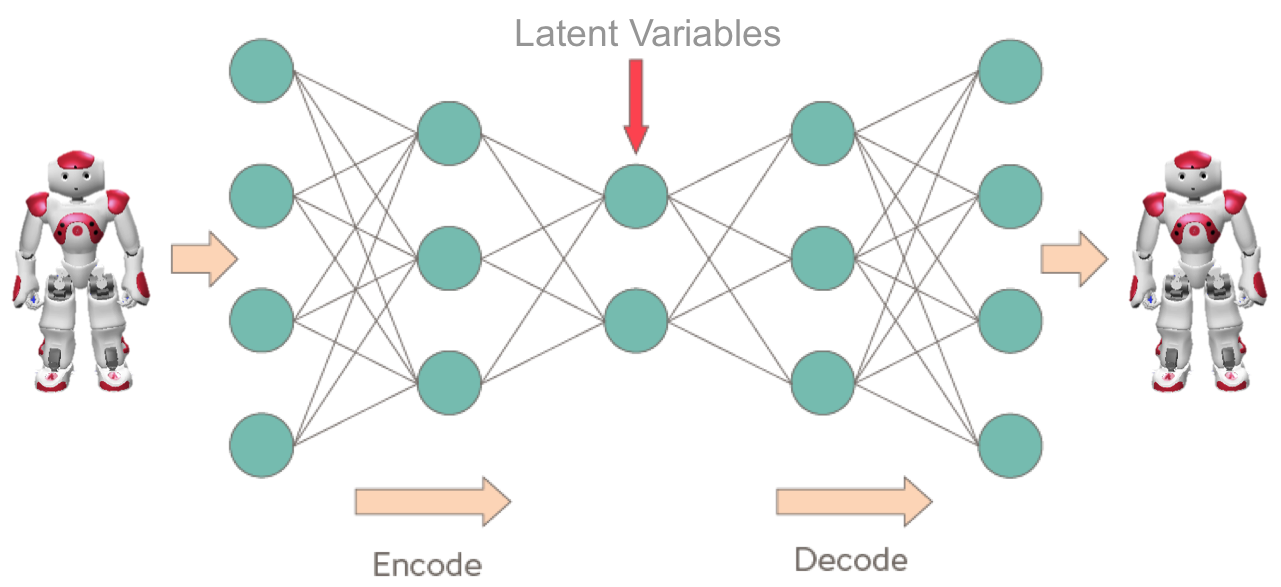}
\caption{Architecture of Variational Autoencoder network. The network is able to faithfully reconstruct the input patterns. }
\label{fig:vae}
\end{center}	
\end{figure}

As it can be seen in \cite{Vitanyi1997}, compression and prediction are closely related fields, and compressors can also be used to generate new data. 

Variational autoencoders, first introduced in \cite{kingma2013auto}, have raised much interest for their capability to produce a variation of the learned input data. 

Their most interesting feature is the capability to autonomously draw the boundaries of a latent space in which the input data can be represented.
Given input data $x$ and calling $p(x)$ the probability distribution of the data, we want to learn the latent variable $z$ with its probability density $p(z)$ so that data can be generated when the values of $z$ are varied:
 
 \begin{equation}
 p(x) = \int p(x|z)p(z)
 \label{eq:gen}
 \end{equation}

The training of the variational autoencoder is based on the variational inference to estimate the distribution $p(x|z)$. This method is often used in Bayesian methodology when you desire to infer a posterior that is difficult to compute. A simpler distribution $q_\lambda(z|x)$ is thus chosen as to minimize the Kullback-Leibler divergence between these two distributions. The variational parameter $\lambda$ is used to refer to a family of distributions and, for a Gaussian family, would represent mean and variance. 
 The divergence is calculated as:

 \begin{equation}
D_{KL}(q_\lambda(z|x)||p(z|x))= \textbf{E}_q[log\frac{q_\lambda(z|x)p(x)}{ p(x,z)}]
 \label{eq:KL}
 \end{equation}

It can be demonstrated that:

 \begin{equation}
log(p(x)) = L^v + D_{KL}(q_\lambda(z|x)||p(z|x))
 \label{eq:L}
 \end{equation}
Since \ref{eq:KL}, to minimize the log(p(x)) it is sufficient to minimize $L^v$. Its value can be calculated as
 \begin{equation}
L^v = -D_{KL}(q(z|x)||p(z))+ \textbf{E}_{q(z|x)}log(p(x|z))
 \label{eq:Lv}
 \end{equation}
Where the first term is $-D_{KL}(q(z|x)||p(z))$ representing the regularization part imposing the distribution of p(z) as similar as possible to $q(z|x)$ while the second part $ \textbf{E}_{q(z|x)}log(p(x|z))$ takes into account a proper reconstruction of the values of $x$. 
 After the training phase aimed at minimizing the value of $log(p(x))$, that is equivalent to maximizing the likelihood, the values of zeta represent the best compression for the input values and variation is the $z$ space corresponds to a variation in the reconstruction of input samples

\section{Creative Robot Dance with variational encoder}
\label{sect:dance}

Throughout this work the expression ``variational encoder'' is used to signify a change of the intended use of variational autoencoders; while the internal structure remains unchanged, latent variables are not used to allow a faithful reconstruction, but rather to introduce a different kind of information that proactively alters the reconstruction, enabling the robot to perform a different set of movements and also change its dancing style according the past performances.

\begin{figure}[htb]
\begin{center}
\includegraphics[width=0.45\textwidth]{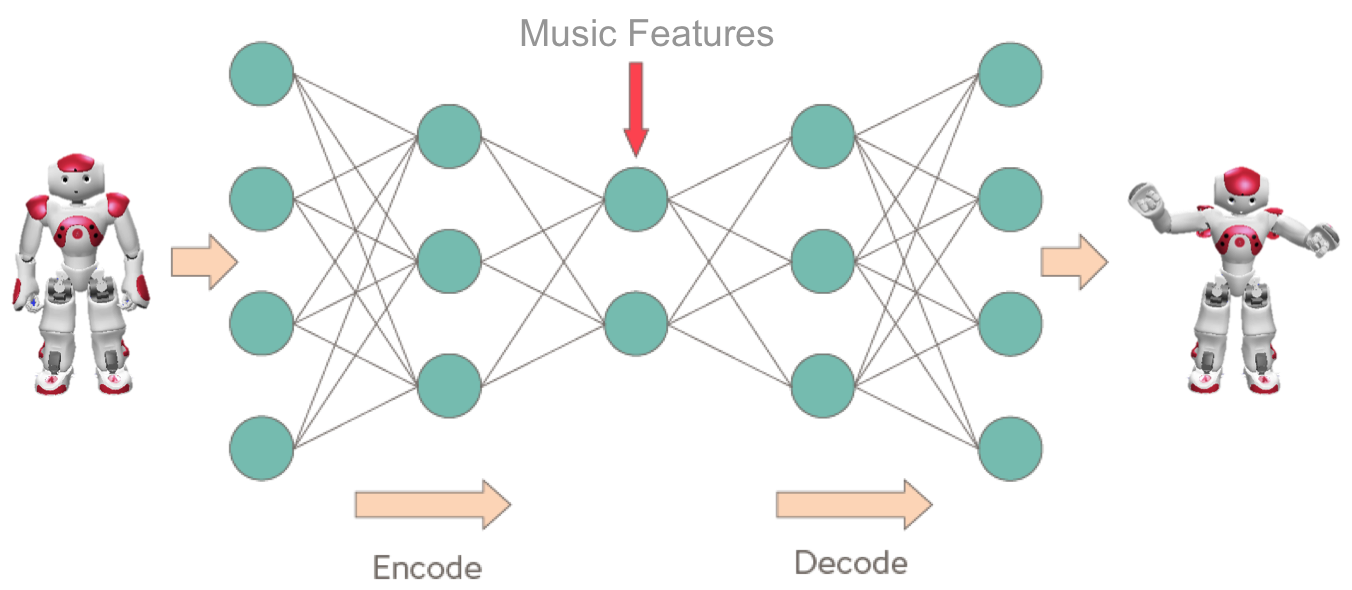}
\caption{Architecture of Variational Encoder network. The network is able to reconstruct sequence of movement that can be varied giving the music features as an additional input. }
\label{fig:vaeM}
\end{center}	
\end{figure}

\begin{figure*}[htb]
  \centering
  \begin{subfigure}[htb]{0.5\textwidth}
    \centering
    \includegraphics[width=0.5\textwidth]{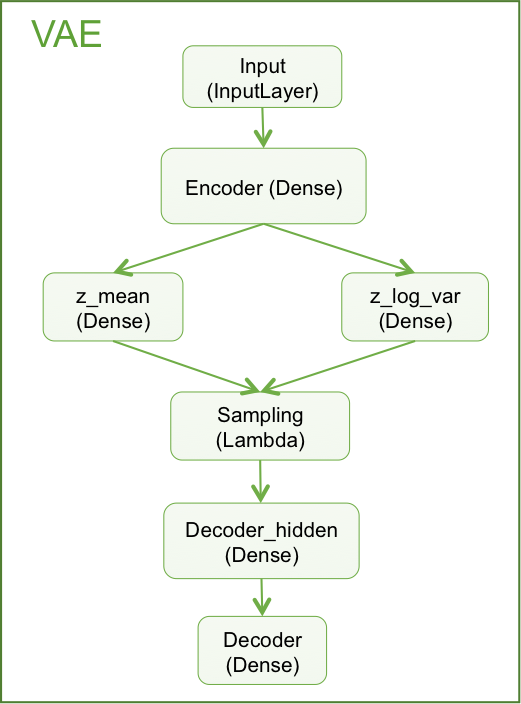}
		\caption{End-to-end model of the variational autoencoder}
		\label{fig:VaeModel}
  \end{subfigure}%
~
  \begin{subfigure}[htb]{0.5\textwidth}
    \centering
    \includegraphics[width=0.5\textwidth]{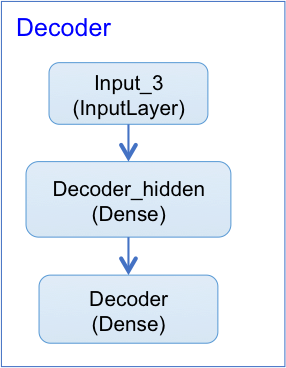}
		\caption{Detailed view of the decoder/predictor}
    \label{fig:PredictModel}
  \end{subfigure}
  \caption{Block diagrams for the variational autoencoder}
\end{figure*}

 The building blocks of a variational encoder are shown in Figure \ref{fig:VaeModel} . It is assumed that the information we deal with can be faithfully approximated with a Gaussian distribution, so that the encoder network can map the input samples into two parameters in a latent space, $z_{mean}$ and $z_{log_{sigma}}$. They uniquely identify a given Gaussian distribution from which randomly sampled points $z$ are then extracted. 
One of our contributions lies in the sampling function: altering the distinctive parameters of the Gaussian curve results in a different output mapping by the decoder network.

\begin{figure*}[htb]
\begin{center}
\includegraphics[width=0.35\textwidth]{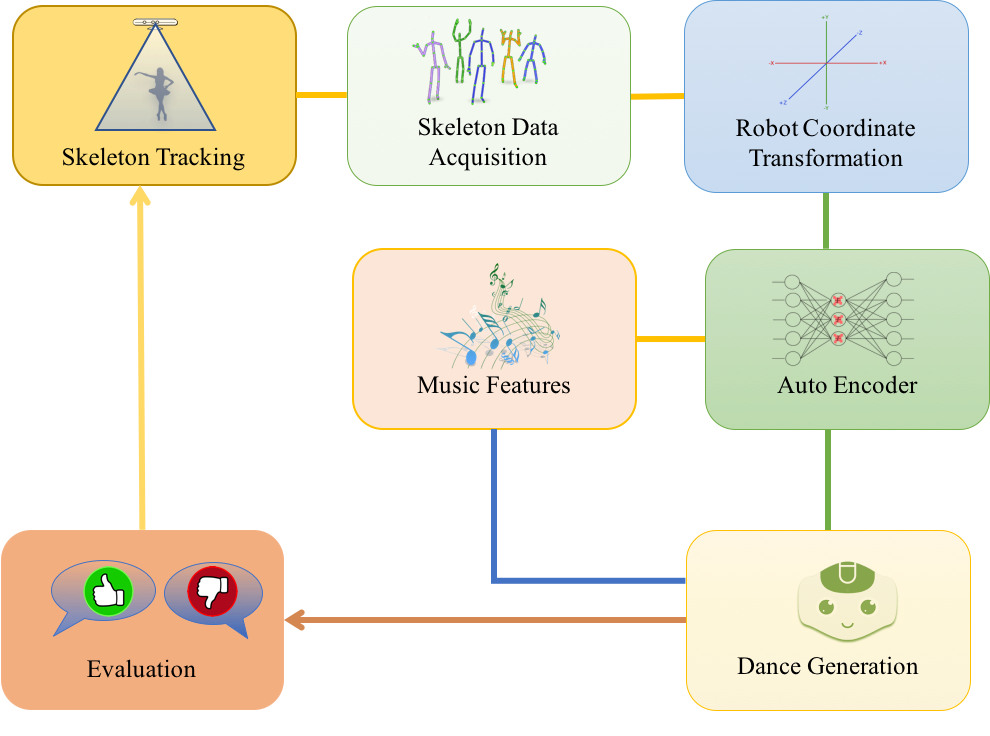}
\hspace{10mm}
\includegraphics[width=0.46\textwidth]{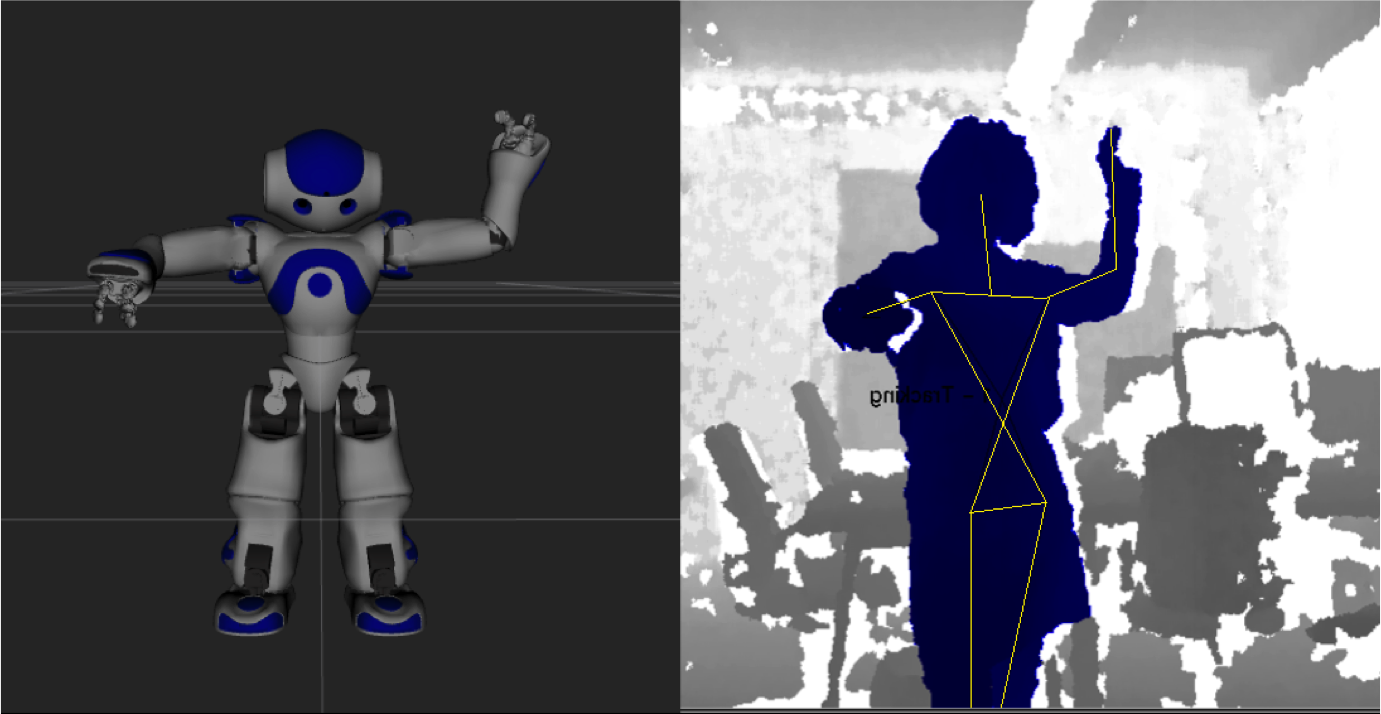}
\caption{On the left side the schema of the learning phase used to obtain a set of dance movements by human demonstration is depicted. On the right there is the screenshot of the skeleton acquisition process during the learning phase and the corresponding posture of the simulated robot.}
\label{fig:LearningSystem}
\end{center}	
\end{figure*}

The parameters of the model are trained taking into account the reconstruction loss that forces the decoded samples to match the initial inputs. Furthermore, we minimize the Kullback-Leibler divergence between the learned latent distribution and the prior distribution, in order to avoid overfitting of the original dataset.

The decoding part, formed the hidden layer of the decoder and the decoder itself is shown in figure \ref{fig:PredictModel} and it is used for the prediction of the output movements. 
In our system, the creation of the robotic dance is based on the three processing phases: processing the sound, learning the movements and the generation of a sequence of movements.

The sound perceptions is implemented extracting some music features that represent the information in the listened music. The generation of the movements is based on the learning phase of the neural network. The execution is the combination of the conceived movements with the perceived music, synchronizing the motions with the rhythm. 
In the following sub sections details of the processes are given.

\subsection{Learning phase}

A basic understanding of the learning process in a human brain is needed when considering the same process in a neural network.
It is believed that, during the learning process, the neural structure is altered by increasing or decreasing the strength of synaptic connections involved in a given activity. Artificial neural networks model this process by adjusting the weighted connections between neurons. Finding a satisfactory configuration may require several iterations which are collectively referred to as ``training''.

In this work, a choreography is built around sequences of movements, that is, sequences of couples of poses. 
The dataset of joint values is partitioned into two subsets: the first one will be used to train the variational autoencoder, the second one for the prediction.

As detailed in previous sections, a correct choice of latent parameters is key to obtain a satisfactory reconstruction of the original input. Gaussian sampling is performed by taking into account both the loudness and the variance of a given music score; the mean value of the curve is the mean of loudnesses, and its standard deviation is the mean of music variances.
A cycle of forward propagation of all the inputs and backward propagation of errors is called ``epoch''. The number of training examples in one forward/backward pass is called ``batch size''. 
In multi-layered networks backward propagation of errors for training is often used in conjunction with an optimization method. 

In this work we used a variant of the stochastic gradient descent (SGD) optimization algorithm called Adadelta, first described in \cite{DBLP:journals/corr/abs-1212-5701}. An extension to a previous  algorithm called Adagrad \cite{Duchi:2011:ASM:1953048.2021068}, it adapts the learning rate to the frequency of parameters; in contrast to the original technique, only a fixed-size history of $w$ squared gradients is considered, instead of the whole set of past gradients.

Let us call $\theta$ the parameters of the training set, $J(\theta)$ the objective function, and $g_{t}$ the gradient of the objective function at time step $t$:

\begin{equation}
g_{t} = \nabla_{\theta} J(\theta)
\end{equation}

To take history values into account, a running average over the gradient is introduced, depending only on the previous average and the current gradient:
\begin{equation}
E[g^{2}]_t = \gamma E[g^{2}]_{t-1} + ( 1 - \gamma ) * g_{t}^2
\end{equation}

where $\gamma = 0.9$.

A SGD update can be described using the following equation:

\begin{equation}
\theta_{t+1} = \theta_t - \eta * g_{t,i} = \theta_t + \Delta\theta_t
\end{equation}

where $\Delta\theta_t$ is the parameter update vector.

It can be demonstrated that the parameter update vector can be rewritten as follows:

\begin{equation}
\Delta\theta_t = - \frac{\eta}{\sqrt{E[g^{2}]_t} + \epsilon} g_{t} = 
- \frac{\eta}{RMS[g]_t} g_{t}
\end{equation}

If the decaying average over squared parameter updates is defined as:

\begin{equation}
E[\Delta \theta^{2}]_t = \gamma E[\Delta \theta^{2}]_{t-1} + ( 1 - \gamma ) * \Delta \theta{t}^2
\end{equation}

the update rule in a way that is not dependent on  the learning rate $\eta$:

\begin{equation}
\theta_{t+1} = \theta_t - \frac{RMS[\Delta\theta]_{t-1}}{RMS[g]_t} g_t
\end{equation}

\subsection{Robotic Dance}

In order to have a dataset of variegated movements, during multiple sessions we have recorded and stored the movements of four users having different experience.
The dataset is composed of both slightly harmonic repetitive sequences and movements with high variability and harmonicity.
The different dance experience and the multiple session of the users provide a diversified dataset since a human usually do not use the same sequences of movements during an improvised dance with music.
The heterogeneity of the dataset is due to the different level of dance experience of the users. In our opinion, all these movements may contribute to creating new unseen movements mixing the learned samples. 


The generation of the movements starts with the extraction of some feature from the music input, in particular, the system extracts the loudness and the variance of the rhythmic sound that will be used as input of the latent space of the variational autoencoder to generate the movements according to the perceived music.


Since the latent space of the network is Gaussian, the values of loudness and variance have to be transformated through the inverse cumulative distribution function of the Gaussian to produce coherent values with the latent space. The transformed value are the real input of the network's latent space and these value lead the generation of the movement related to the perceived music.

Our idea is to allow the robot to execute one motion according the beat. Given the position of the beat and the interval between two consecutive beats, the system can execute one movement for each detected interval.
The network, using as an input the intensity and variance of the music interval outputs a configuration of joints that represents one single movement. The numbers of movements that the network predicts is function of the number of processed music features. 

The final part of the system focuses on the execution of the movements by the robot in synchrony with the music.
The execution of the dance is made up combining the movements predicted by the network and the features of the audio signal to keeping time, in fact, we use the information of the beat position and beat interval to regulate the duration of each movement.
\\
 The proposed system is flexible and can adapt to different music genres, in fact, whatever type of rhythmic music is provided as an input, the system can generate sequences of movements.
Moreover, it is possible to continue to train the network with others music genres, adding new movements captured in different learning sessions with human dancers.

\section{Experimental Results and Discussion}

In the following subsections we describe the experimental setup, the results and a brief discussion about evaluation issues of the system.

\subsection{Dance Movement Acquisition}
\label{sect:dataset}


To collect the dataset of the movements, we employed the Microsoft RGB-D\footnote{Red Green Blue plus Depth}  Kinect camera to track the improvised movements of the human dancers. We used the Kinect camera since it is non-invasive and cheap compared to other motion capture systems, avoiding to use high precision capture devices. Using a Kinect camera the dancer does need to wear any device and he can act in a natural manner; moreover, considering that the robot cannot reproduce all the possible human posture due to its structural limitation  more precision does not resolve such problem.


The Kinect includes a RGB camera and four microphones to capture sounds, an infrared (IR) emitter and an IR depth sensor to capture the depth map of each image. The collected information enables the extraction of the spatial position of the dancer's joints in a non-invasive way. 
The advantage in employing the kinect is that any person in a room can be recorded as a teacher for the dance without the any preparation and without setting the connections that are required for the body sensors.

The use of Kinect camera allows capturing graceful and pleasant movements displayed in front of the acquisition system and the good sampling frequency, around thirty frame per second, allows to maintain a high correlation between a motion and the next one.

To interface the acquisition device with the NAO robot, that is used to reproduce natural motion, we used ROS (Robotics Operating System)\footnote{{http://wiki.ros.org/it}}. It is an open-source operating system for robots that provides a layered structure to communicate in a peer-to-peer topology between server and hosts. ROS allows the user to connect different hosts at runtime, managing messages among processes, sensors and actuators.

Through the Kinect sensor it is possible to extract the skeleton data of a human dancer, obtaining in real-time the list of the 15 joint position of the human body along the three axis (x,y,z). 
The skeleton information cannot be directly used for the robot positions since the coordinate system are different and the robot has several limitations if compared to human movements.
Hence, the skeleton data extracted should be transformed to change the coordinate system to perform the movements in the robot. To track the position of the arms and the head during the motion of the dancer we have used the ROS package \emph{skeleton markers}\footnote{wiki.ros.org/skeletonmarkers}. In \cite{rodriguez2014humanizing} is described a system, based on the same technologies, to set up a robust teleoperation system.
In the current implementation a stronger focus has been given to the movement of the upper body part not to compromise the stability of the robot during the dance.


The dataset of the movements has been created by observing the dance of four different people; \emph{User 1} and \emph{User 2} have a limited competence in the dance field. \emph{User 3} owns an expertise in dance, she is not a professional but often performs dance and has a good sense of rhythm. \emph{User 4} is a professional in couple dancing, highly skilled to follow the rhythm. 
They were asked to execute spontaneous motions while listening to different songs, while the RGB-D camera of the Microsoft Kinect v1 acquired the sequence of their movements and saved photo shots.

The system tracks and samples the movements of the human dancers as soon as music beats are detected. 
After the conversion in the robot coordinate system, the transformed joint positions are stored to be subsequently used during the training phase. Since the robot has less degrees of freedom than human beings, some complex movements appear to be truncated if compared with the original ones. For example, the robot is not able to perform sinusoidal movements with the arm or push forward its shoulder.

\begin{table*}[htb]
\centering
\caption{Evaluation of Robot movements versus the learning epochs}
\label{Movements}
\begin{tabular}{M{7cm}|N{10cm}}
\begin{center}\textit{\textbf{Poses}}\end{center} & \begin{center}\textit{\textbf{Comments}}\end{center} \\
 \hline
 \vspace*{0.5cm}
 \includegraphics[height=2.5cm]{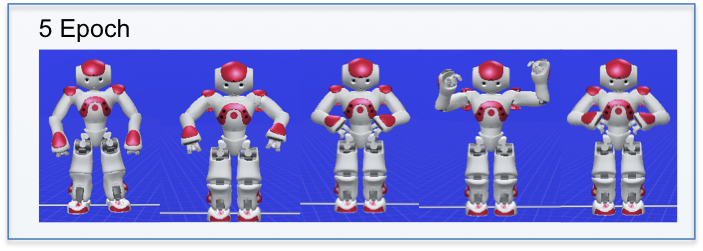}  & The robot basically repeats the same simple movements. \\
 \includegraphics[height=2.5cm,width=7cm]{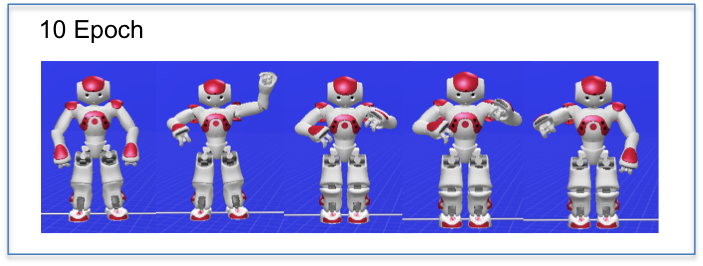} & There is some more variability. The left arm is risen independently.\\
 \includegraphics[height=2.5cm,width=6.95cm]{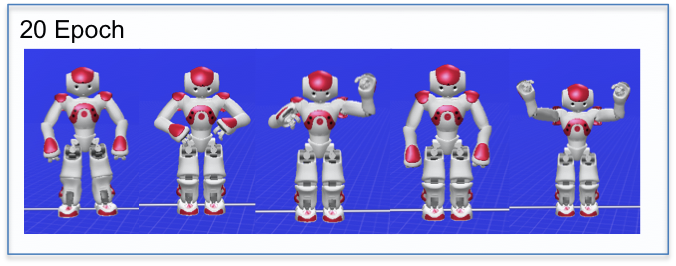} & Much more variability. Movements become more complex and do not seem to follow a pre-recorded pattern. \\
 \includegraphics[height=2.5cm,width=6.95cm]{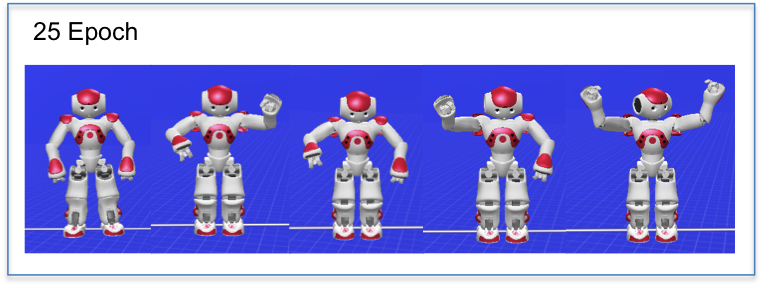} & Maximum variability. Uses both new and old poses. Even the head starts moving.	
\end{tabular}
\end{table*}


\subsection{Sound Processing}

The Essentia Library \cite{bogdanov2013essentia} has been used to identify music features that will be used to generate movements, enabling the robot to dance in synchrony with the music.

The following features are extracted: Beats location, Beat Per Minute (BPM), beat interval, variance and loudness.

Beat locations give the timestamps where the rhythm falls, whereas beat intervals indicate the interval of time between two consecutive beats; both are used to compute the intensity and the variation within two subsequent musical beats and to synchronize the movements with the music.

From the beat positions we calculate the loudness and the variance within the frame between two consecutive beats that are used in the successive steps to generate movements by mean the network.

\subsection{Dancing Experiments}

Robotic dance information has been acquired using a set of custom Python scripts.
The autoencoder has been implemented using the Keras\footnote{{http://keras.io/}} \cite{chollet2015} open source framework for rapid prototyping of deep networks.\\
To train the model four datasets are needed:

\begin{itemize} 
\item Two sets $P_1$, $P_2$ containing $k$ joint poses. 
\item Music variances $V$
\item Music loudnesses $L$
\end{itemize}

The mean of values in $V$, $V_m$, and the mean of values in $L$, $L_m$, are calculated; they will be used to create the Gaussian distribution used to sample latent features.

Successive joint poses in $P_1$ and $P_2$ are coupled to form two sets $M_1$ and $M_2$ containing $k/2$ movements. The set $M_2$ is $M_1$ forward shifted of 1 time unit. Movements in $M_1$ and $M_2$ will be supplied as input and as expected output, respectively, to train the encoder.

When training process has converged new movements can be generated providing values of variance and loudness values to the latent units of the network. If the values are extracted from the musical piece, the robot can improvise a dance following the features of the listened song.

Table \ref{tab:variance_joints} shows the variance of generated joint values as the requested number of epochs increases up to the 25th, which is the last we have considered, as seen in \ref{sect:VE}, as further runs do not produce an appreciable decrease in the error function.

\begin{table*}[htb]
\centering
\caption{Variance of the joints}\label{tab:variance_joints}
\label{VarianceofJoints}
\begin{tabular}{l|lllllll}
\textit{\textbf{Names of Joints}} & \textit{\textbf{Epoch 1}} & \textit{\textbf{Epoch 5}} & \textit{\textbf{Epoch 10}} & \textit{\textbf{Epoch 15}} & \textit{\textbf{Epoch 20}} & \textit{\textbf{Epoch 25}} \\
\hline
LElbowRoll   & 0.0584 & 0.8002 & 0.1806 & 0.1651 & 0.1522 & 0.3280  \\
RElbowRoll   & 0.0390 & 1.0252 & 0.4269 & 0.1781 & 0.1924 & 0.3049 \\
LElbowYaw   & 0.2596 & 0.2529 & 0.2095 & 0.1018 & 0.6654 & 0.3256  \\
RElbowYaw   & 0.2219 & 0.3678 & 0.2021 & 0.3635 & 0.1965 & 0.1898  \\
LShoulderRoll & 0.1045 & 0.1724 & 0.2383 & 0.0376 & 0.3827 & 0.4827  \\
RShoulderRoll & 0.0253 & 0.1207 & 0.2563 & 0.3597 & 0.5258 & 0.2854  \\
LShoulderPitch & 0.3577 & 0.1392 & 0.1505 & 0.1388 & 0.3065 & 0.7498  \\
RShoulderPitch & 0.1993 & 0.1547 & 0.2927 & 1.2272 & 1.1995 & 1.4630 \\
HeadYaw    & 0.0416 & 0.0980 & 0.0539 & 0.0306 & 0.0176 & 0.0152 \\
HeadPitch   & 0.0001 & 0.0002 & 0.0003 & 0.0000 & 0.0001 & 0.0002 
\end{tabular}
\end{table*}

The variance of joint configurations measures how far robot movements deviate from the mean; higher values of variance may thus be used to signify an increased creativity in motion sequence generation. This insight is substantiated if we consider some example movements generated at different epochs, as shown in the table \ref{Movements}.
%
%
%

While at epoch 1 movement sequences are quite repetitive, they become more and more harmonic as the neural network continues the training. At epoch 25 a remarkable distinctiveness can be detected.

\begin{figure*}[tb]
  \centering
  \begin{subfigure}[htb]{0.5\textwidth}
    \centering
    \includegraphics[width=\textwidth]{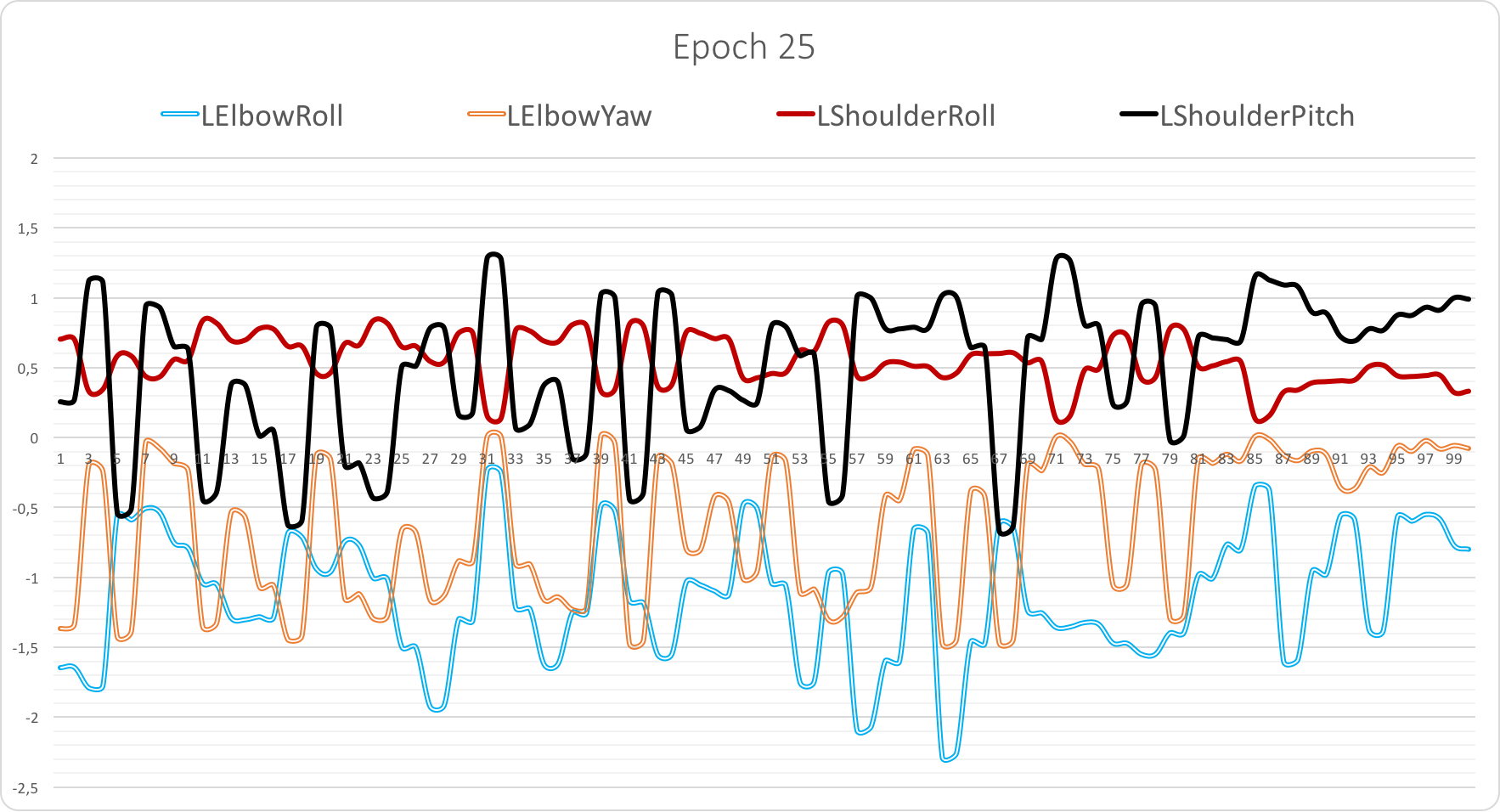}
		\caption{Left arm chain of the robot}
		\label{fig:L4joint}
  \end{subfigure}%
~
  \begin{subfigure}[htb]{0.5\textwidth}
    \centering
    \includegraphics[width=\textwidth]{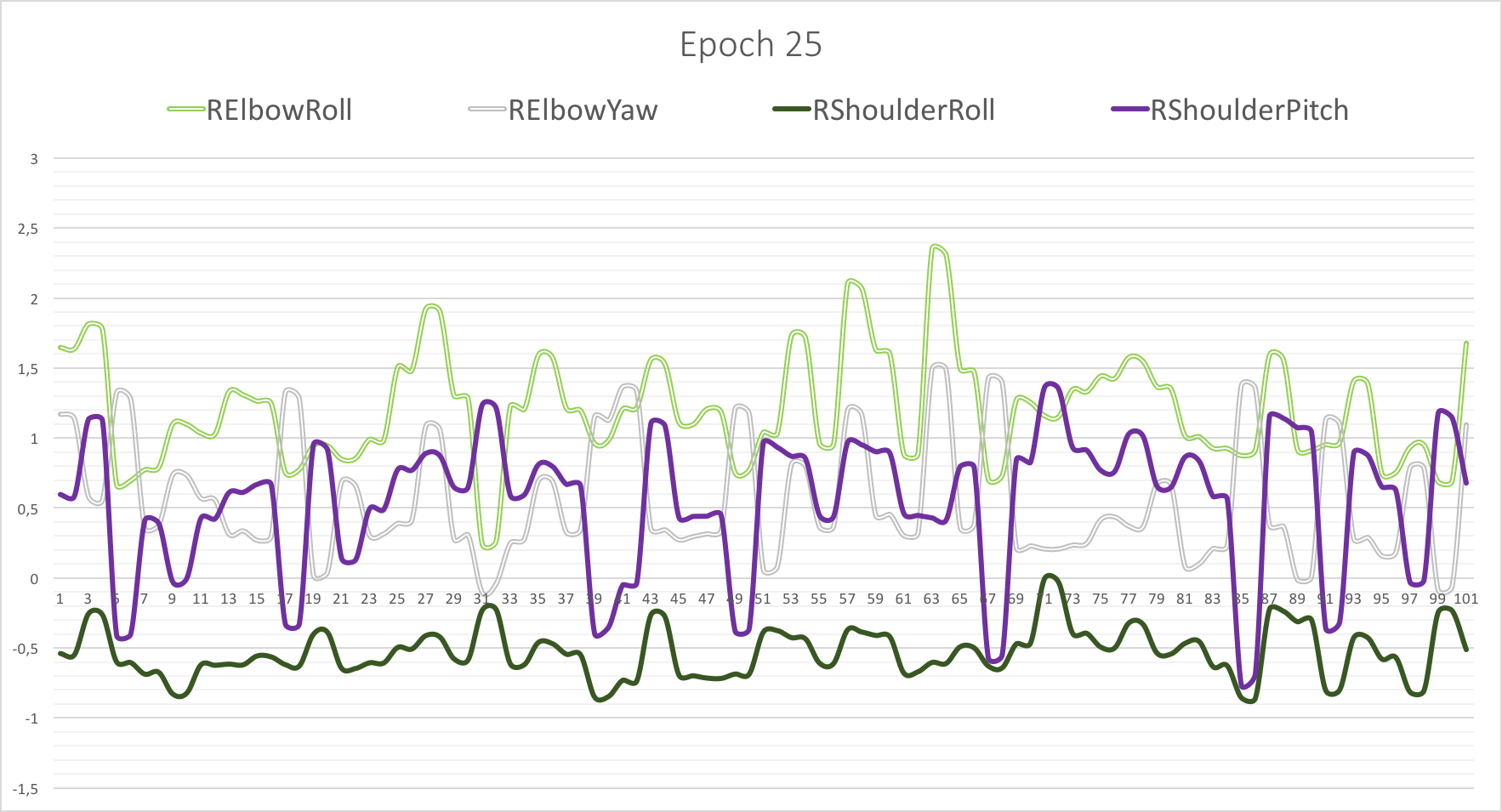}
	\caption{Right arm chain of the robot}
    \label{fig:R4joint}
  \end{subfigure}
  \caption{Progress of the joints values during a dance session}
    \label{fig:joints}
\end{figure*}

\begin{table}[htb]
\centering
\caption{Mean of the variance of the joint value during different epochs }
\label{MeanVariance}
\begin{tabular}{l|l}
     & \textit{\textbf{Variance Mean}} \\
     \hline
\textit{\textbf{Epoch 1}} & 0.1307   \\
\textit{\textbf{Epoch 5 }} & 0.3131   \\
\textit{\textbf{Epoch 10}} & 0.2011   \\
\textit{\textbf{Epoch 15}} & 0.2603   \\
\textit{\textbf{Epoch 20}} & 0.3639   \\
\textit{\textbf{Epoch 25}} & 0.4145   
\end{tabular}
\end{table}

We have also computed mean of the variance of the joint value during different epochs.
An interesting consideration to be taken into account is that the trend of variance stalls between 10 and 20 epochs, and then reaches a maximum at epoch 25.
In Figure \ref{fig:joints} are shown the plot of the angle values (in radiant) of the right and left joint in the should and elbow. The value are referred to epoch 25 and show a rich set of movements learned at the final steps.

In our opinion the evaluation of the output of artificial systems is a key issue for well founded computational creativity. The implementation of the dance with neural encoders tends to focus on a restricted set of movements and to repeat the same patterns. From this point of view the variance of the movements is a key parameter indicating that a large set of movements has been learned and the dance resembles human movements.
Furthermore an extern evaluation from people staring at the dance performance can be used to select the most adapt movements for a robotic dance.
In previous work \cite{manfreICRC17}, we used a clustering approach to define groups collecting similar dance actions. Within the cluster, each movement has an evaluation score initially set to 100. The centroid of the cluster is the first representative of the group used to determine the dance creation as previously described. If a user evaluates the performance as negative, then the scores of the movements belonging to the sequence are lowered. After hundreds of performances, when a movement has a score below a given threshold (e.g. less than 50), the system searches for the substitute with the highest score in the same cluster. The positive evaluation causes an increment of the related scores of involved movements. The judgment of an expert evaluator provides another evaluation mechanism that has a strong influence on dance execution. In fact, the expert could indicate inadequate a single movement or a short sequence and directly causes an inhibition  (i.e. the score is equal to 0).

In \cite{augello2016creation} we report the evaluation results of various live performances with heterogeneous audiences. Even we have no rigorous experimental evidence the results seem to demonstrate that variability of movement is an important factor for a positive evaluation.
A video showing the progress during the training can be found at: \url{www.youtube.com/watch?v=arnaoFPjfUc}

\section{Conclusions}

In this work, we proposed a deep learning approach to induce a computational creativity behavior in a dancing robot. In particular we used a variational encoder that allows mapping input patterns in a latent space. 

The encoder has been trained with a set of movements captured from differently skilled dancers.  The generation is obtained by injecting the representation of the listened music in the latent space of the encoder network. 
As a result, the robot is able to improvise dancing movements according to the listened music even if it has not been previously presented in the learning phase.
\balance

%
\label{sect:bib}
\bibliographystyle{iccc}

\end{document}